\documentclass[10pt,twocolumn,letterpaper]{article}

\usepackage[pagenumbers]{iccv} 

\usepackage[noend]{algpseudocode}
\usepackage{algorithm}
\usepackage{graphicx}
\usepackage{comment}
\usepackage{amsmath}

\usepackage{epsfig}
\usepackage{multirow}
\usepackage{boldline}
\usepackage{array}

\definecolor{iccvblue}{rgb}{0.21,0.49,0.74}
\usepackage[pagebackref,breaklinks,colorlinks,allcolors=iccvblue]{hyperref}

\title{VCMamba: Bridging Convolutions with Multi-Directional Mamba for Efficient Visual Representation}

\author{Mustafa Munir* \quad Alex Zhang* \quad Radu Marculescu \\
The University of Texas at Austin\\
{\tt\small \{mmunir, alex.zhang, radum\}@utexas.edu} \\
}

\begin{document}
\maketitle

\def\thefootnote{*}\footnotetext{Equal contribution}

\begin{abstract}
Recent advances in Vision Transformers (ViTs) and State Space Models (SSMs) have challenged the dominance of Convolutional Neural Networks (CNNs) in computer vision. ViTs excel at capturing global context, and SSMs like Mamba offer linear complexity for long sequences, yet they do not capture fine-grained local features as effectively as CNNs. Conversely, CNNs possess strong inductive biases for local features but lack the global reasoning capabilities of transformers and Mamba. To bridge this gap, we introduce \textit{VCMamba}, a novel vision backbone that integrates the strengths of CNNs and multi-directional Mamba SSMs. VCMamba employs a convolutional stem and a hierarchical structure with convolutional blocks in its early stages to extract rich local features. These convolutional blocks are then processed by later stages incorporating multi-directional Mamba blocks designed to efficiently model long-range dependencies and global context. This hybrid design allows for superior feature representation while maintaining linear complexity with respect to image resolution. We demonstrate VCMamba's effectiveness through extensive experiments on ImageNet-1K classification and ADE20K semantic segmentation. Our VCMamba-B achieves 82.6\% top-1 accuracy on ImageNet-1K, surpassing PlainMamba-L3 by 0.3\% with 37\% fewer parameters, and outperforming Vision GNN-B by 0.3\% with 64\% fewer parameters. Furthermore, VCMamba-B obtains 47.1 mIoU on ADE20K, exceeding EfficientFormer-L7 by 2.0 mIoU while utilizing 62\% fewer parameters. Code is available at \url{https://github.com/Wertyuui345/VCMamba}.

\end{abstract}

\vspace{-6mm}


\section{Introduction}
\label{sec:intro}

The landscape of deep learning vision architectures has been predominantly shaped by Convolutional Neural Networks (CNNs) \cite{lecun1998gradient, Alexnet2012, resnet, convnext}. Their inherent inductive biases, such as locality and weight sharing, enable efficient learning of hierarchical features. However, the fixed receptive fields of convolutions can limit their ability to capture long-range dependencies effectively. Vision Transformers (ViTs) \cite{ViT} emerged as a powerful alternative, treating images as sequences of patches and leveraging self-attention mechanisms to model global relationships \cite{vaswani2017attention}, often achieving superior performance, albeit with quadratic complexity with respect to the number of patches.

More recently, State Space Models (SSMs) have garnered attention as a promising architecture. Mamba \cite{gu2023mamba}, a notable SSM variant, introduces a selection mechanism and a hardware-aware algorithm, enabling linear-time sequence modeling and strong performance on language tasks. This has spurred interest in adapting SSMs for vision, leading to models like VMamba \cite{liu2024vmamba}, Vision Mamba (Vim) \cite{vim}, and PlainMamba \cite{yang2024plainmamba}. VMamba adapts one-dimensional SSMs to two-dimensional visual data using a 2D Selective Scan Module \cite{liu2024vmamba}. Vim utilizes bidirectional Mamba blocks for visual representation \cite{vim}, while PlainMamba proposes a simple, non-hierarchical SSM with specific 2D scanning adaptations for visual recognition \cite{yang2024plainmamba}. These models demonstrate the potential of SSMs in vision, offering an attractive balance of performance and efficiency, particularly for high-resolution inputs \cite{yang2024plainmamba, liu2024vmamba, vim}.

Despite these advances, many contemporary vision SSMs, similar to ViTs, primarily rely on patch embeddings that might not fully capitalize on the rich, spatially dense local features that CNNs excel at capturing in the early stages of visual processing. While PlainMamba focuses on a non-hierarchical SSM structure \cite{yang2024plainmamba}, and other vision SSMs explore different scanning strategies or bidirectional mechanisms \cite{liu2024vmamba, vim}, there remains an opportunity to create a vision backbone that explicitly fuses the robust local feature extraction of hierarchical convolutional stages with the efficient global modeling of advanced multi-directional Mamba modules in the later stages. Standard non-overlapping patch embeddings can lead to information loss, especially for fine-grained details crucial for dense prediction tasks, a challenge a hierarchical convolutional structure for higher resolutions stages can mitigate.

To bridge this gap, we introduce \textit{VCMamba}, a novel vision backbone that integrates the strengths of CNNs and multi-directional Mamba SSMs. VCMamba employs a convolutional stem and a hierarchical structure with convolutional blocks (realized as feed-forward networks with 1$\times$1 and 3$\times$3 convolutions) in its early and intermediate stages, enabling the extraction of rich, multi-scale local features. These features are then processed by later stages incorporating multi-directional Mamba blocks, specifically leveraging the 4-way scanning mechanism detailed in \cite{yang2024plainmamba}, to efficiently model long-range dependencies and global context. This hybrid, hierarchical design allows for superior feature representation while maintaining linear complexity with respect to image resolution in its Mamba stages, a key advantage for high-resolution image processing \cite{yang2024plainmamba, liu2024vmamba, vim}.

We demonstrate VCMamba's effectiveness through extensive experiments on ImageNet-1K classification \cite{imagenet1k} and ADE20K semantic segmentation \cite{ADE20K}. VCMamba-B achieves 82.6\% top-1 accuracy on ImageNet-1K, surpassing PlainMamba-L3 (82.3\% Acc \cite{yang2024plainmamba}) by 0.3\% with 37\% fewer parameters, and outperforming Vision GNN-B (82.3\% Acc \cite{Vision_GNN}) by 0.3\% while using 64\% fewer parameters. Furthermore, VCMamba-B obtains 47.1 mean intersection over union (mIoU) on ADE20K semantic segmentation, exceeding EfficientFormer-L7 (45.1 mIoU \cite{EfficientFormer}) by 2.0 mIoU while utilizing 62\% fewer parameters. These results showcase VCMamba's strong performance and efficiency, particularly in tasks benefiting from its hybrid feature extraction capabilities. Our contributions are:
\begin{itemize}
    \item We propose VCMamba, a novel hierarchical vision architecture that effectively combines multi-stage convolutional feature extraction with multi-directional Mamba SSMs for efficient and powerful global context modeling.
    \item We propose several VCMamba variants and demonstrate their superior performance against leading CNN, ViT, vision GNNs and vision SSM architectures.
    \item We show that our hybrid and hierarchical approach leads to strong performance in both image classification and semantic segmentation tasks, retaining the efficiency benefits of Mamba for high-resolution inputs.
\end{itemize}

The paper is organized as follows. Section \ref{sec:related_work} covers related work. Section \ref{sec:preliminaries} explains preliminary information on state space models. Section \ref{sec:methodology} describes our hierarchical feature extraction, multi-directional mamba blocks, and our VCMamba architecture. Section \ref{sec:experiments} describes our experimental setup and results for ImageNet-1K image classification and ADE20K semantic segmentation. Lastly, Section \ref{sec:conclusion} summarizes our main contributions.

\section{Related Work}
\label{sec:related_work}

\subsection{Vision Architectures}

\noindent
\textbf{Convolutional Neural Networks (CNNs)} have long been the dominant architecture in computer vision \cite{lecun1998gradient, Alexnet2012, resnet, convnext, VGGNet}. Their success stems from inherent inductive biases like locality and weight sharing, which allow for efficient learning of hierarchical visual features. Architectures such as ResNet \cite{resnet}, EfficientNet \cite{tan2019efficientnet}, and ConvNeXt \cite{convnext} have consistently pushed performance boundaries. For mobile applications, lightweight CNNs like MobileNet \cite{MobileNet, Mobilenetv2} introduced efficient operations like depthwise separable convolutions. Recent works \cite{munir2025rapidnet, wang2024repvit} have further explored enhancing CNNs for mobile vision by using techniques to expand receptive fields efficiently. While CNNs excel at local feature extraction, their fixed receptive fields can be a limitation for capturing global, long-range dependencies compared to attention-based models.

\noindent
\textbf{Vision Transformers (ViTs)} marked a paradigm shift by applying the Transformer architecture \cite{vaswani2017attention}, originally designed for natural language processing, to image data. ViTs \cite{ViT} treat images as sequences of patches and use self-attention to model global relationships between them, often achieving state-of-the-art results \cite{Deit, liu2021swin, wang2021pyramid}. However, the quadratic complexity of self-attention with respect to the number of patches (image resolution) poses significant computational challenges for high-resolution inputs and dense prediction tasks. Efforts to mitigate this include MobileViT \cite{MobileViT} and MobileViTv2 \cite{MobileViTv2}, which aim to create more efficient ViT variants for mobile devices, often by incorporating convolutional principles. 

\noindent
\textbf{Vision Graph Neural Networks (ViGs)} offer another perspective by modeling images as graphs of interconnected patches or nodes \cite{Vision_GNN}. ViG \cite{Vision_GNN} was an early proponent of using GNNs as a general vision backbone through leveraging K-Nearest Neighbors to connect similar nodes in the graph. For mobile applications, MobileViG \cite{MobileViG} introduced a static graph-based connection mechanism. Further optimizations like GreedyViG \cite{greedyvig2024}, WiGNet \cite{spadaro2025wignet}, and ClusterViG \cite{parikh2025clustervig} focused on introducing new dynamic and efficient graph construction algorithms. While ViGs provide a flexible way to model relationships, the graph construction and propagation steps can introduce computational overhead.

\subsection{State Space Models (SSMs) in Vision}
State Space Models (SSMs) have recently emerged as a compelling alternative for sequence modeling, offering linear complexity with sequence length. Mamba \cite{gu2023mamba}, a prominent SSM, introduced a selective scan mechanism that enables efficient, input-dependent processing and has shown strong performance in language modeling. This success in language modeling has catalyzed significant interest in adapting SSMs, particularly Mamba, for computer vision tasks \cite{visionmamba_survey, liu2024vmamba, vim}. Initial efforts to translate Mamba's capabilities to the visual domain include:

\begin{itemize}
    \item \textbf{VMamba} \cite{liu2024vmamba} proposes the 2D Selective Scan Module (SS2D) to facilitate 2D spatial awareness for SSMs, converting images into ordered patch sequences processed by four-way scanning.
    \item \textbf{Vim} \cite{vim} introduces a vision backbone employing bidirectional Mamba blocks and positional embeddings.
    \item \textbf{PlainMamba} \cite{yang2024plainmamba} focuses on a non-hierarchical Mamba architecture for vision, introducing Continuous 2D Scanning to better adapt Mamba's selective scan to 2D images.
\end{itemize}

Other recent research directions include domain-specific adaptations like VideoMamba \cite{li2024videomamba}, specialized tasks such as quality assessment with QMamba \cite{guan2024qmamba}, and efficiency enhancements like post-training quantization in PTQ4VM \cite{cho2024ptq4vm}. Concurrently, analytical works like MambaOut \cite{yu2024mambaout} and \cite{han2024demystify} critically examine the role and efficacy of Mamba components in the visual domain. These models underscore the potential of SSMs to provide an efficient and powerful backbone for various vision tasks, though active research continues to optimize their application to 2D data.


\subsection{Hybrid Architectures}
The complementary strengths of different architectures have led to various hybrid models. CNN-ViT hybrids like CoAtNet \cite{dai2021coatnet} and MobileFormer \cite{MobileFormer} aim to combine CNNs' local feature extraction with ViTs' global context modeling. EfficientFormer \cite{EfficientFormer, EfficientFormerv2} further optimizes this fusion for speed and efficiency on mobile devices. MambaVision \cite{hatamizadeh2024mambavision} represents a recent Mamba-Transformer hybrid that integrates Transformer blocks into its later stages to improve the capture of long-range spatial dependencies.

Our proposed VCMamba contributes to this line of work by creating a distinct CNN-SSM hybrid. Unlike MambaVision which integrates Transformer blocks with Mamba, VCMamba employs a hierarchical structure with convolutional blocks in its early and intermediate stages for robust multi-scale feature extraction, subsequently transitioning to multi-directional Mamba blocks in its later stages for efficient global modeling. Our approach aims to harness the local feature richness from CNNs and the sequential modeling power of Mamba without relying on Transformer blocks.


\section{Preliminaries: State Space Models (Mamba)}
\label{sec:preliminaries}


State Space Models (SSMs) describe systems via state variables. A continuous linear SSM is defined by:


\begin{equation}
    h'(t) = \mathbf{A}h(t) + \mathbf{B}x(t), \quad y(t) = \mathbf{C}h(t) + \mathbf{D}x(t)
\end{equation}


where $x(t)$ is the input, $h(t)$ is the latent state, $y(t)$ is the output, and $(\mathbf{A}, \mathbf{B}, \mathbf{C}, \mathbf{D})$ are system matrices. For deep learning applications, these are discretized using a timescale parameter $\Delta$ to transform A and B into their discrete counterparts $\overline{\mathbf{A}}$ and $\overline{\mathbf{B}}$. Mamba \cite{gu2023mamba} significantly enhances traditional SSMs by making the parameters $\overline{\mathbf{B}}$, $\mathbf{C}$, and $\Delta$, input-dependent (selective). This allows the model to dynamically modulate its behavior based on the current token, selectively propagating or forgetting information along the sequence. The discretized recurrence is:

\begin{equation}
    h_k = \overline{\mathbf{A}}_k h_{k-1} + \overline{\mathbf{B}}_k x_k, \quad y_k = \mathbf{C}_k h_k + \mathbf{D}_k x_k
\label{eq:mamba_recurrence_prelim}
\end{equation}


where the subscript $k$ indicates the input-dependent nature of the parameters in Mamba \cite{gu2023mamba}. Mamba employs a hardware-efficient parallel scan algorithm for training and inference, achieving linear complexity.

Adapting Mamba for 2D visual data, as in PlainMamba \cite{yang2024plainmamba}, involves flattening image patches into a 1D sequence and then applying the selective scan. To capture 2D spatial context, PlainMamba employs techniques like Continuous 2D Scanning, which processes visual tokens in multiple (e.g., four) pre-defined orders ensuring spatial adjacency, and Direction-Aware Updating, where learnable parameters $\Theta_k$ representing scan directions are incorporated into the SSM's update rule. For instance, the update to the hidden state $h_{k,i}$ for the $k$-th scan direction and $i$-th token can be augmented as:


\begin{equation}
    h_{k,i} = \overline{\mathbf{A}}_{i}h_{k,i-1} + (\overline{\mathbf{B}}_{i} + \overline{\mathbf{\Theta}}_{k,i})x_{i}
\label{eq:plainmamba_direction_aware}
\end{equation}

The outputs from these multiple directional scans are then typically aggregated. This multi-directional approach is crucial for VCMamba's later stages.

\section{VCMamba Architecture}
\label{sec:methodology}

\begin{figure*}[t]
    \centering
    \includegraphics[width=0.7\linewidth]{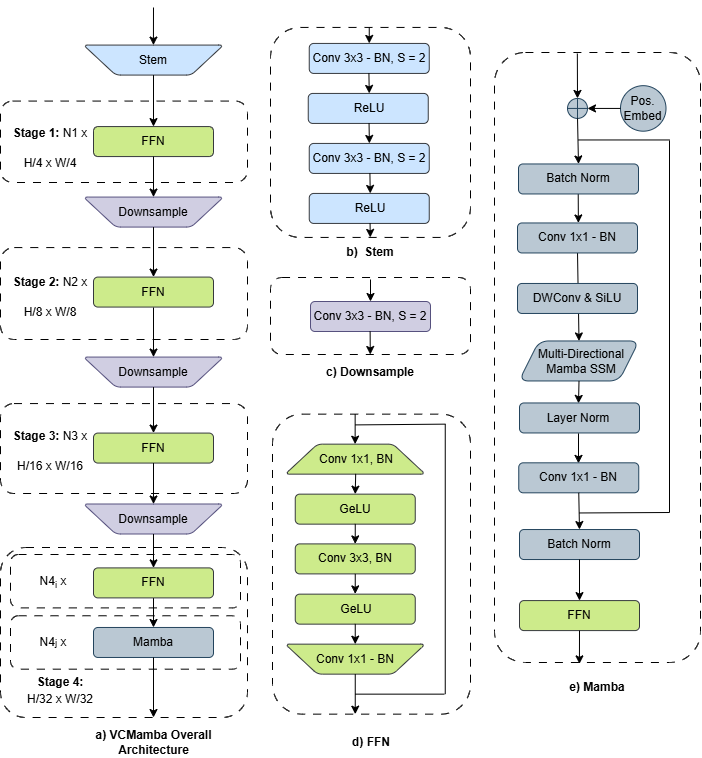}
    \caption{ Overall architecture of VCMamba. An input image is passed through a convolutional head and four downsampling stages. The first three stages consist of FFN convolutional blocks only. The last stage contains several Mamba and FFN blocks.}
    \label{fig:overall_arch}
\end{figure*}

We design VCMamba as a hierarchical vision backbone that integrates the robust local feature extraction capabilities of CNNs in its early stages with the efficient global context modeling of multi-directional Mamba SSMs in its later stages. This hybrid approach allows VCMamba to effectively process visual information at multiple scales while maintaining computational efficiency. The overall architecture, depicted in Figure \ref{fig:overall_arch}, consists of a convolutional stem followed by four stages of feature extraction blocks, with downsampling layers between stages to create a feature pyramid \cite{Vision_GNN, wang2021pyramid}.

\subsection{Hierarchical Feature Extraction}
\label{ssec:stem_and_stages}

The VCMamba architecture initiates processing with a convolutional stem (Fig. \ref{fig:overall_arch}(b)), a common strategy in modern vision backbones \cite{EfficientFormer, convnext} for initial feature extraction and spatial downsampling. This stem comprises two sequential $3 \times 3$ convolutional layers, each employing a stride of 2. Batch Normalization \cite{BatchNorm} and ReLU activation functions \cite{ReLU} follow each convolution. Collectively, the stem downsamples the input image by a factor of 4, efficiently generating low-level feature representations at a reduced spatial resolution for subsequent stages.

After the stem, VCMamba adopts a four-stage hierarchical structure (Fig.\ref{fig:overall_arch} (a)) to progressively refine features and build multi-scale representations \cite{resnet, liu2021swin, munir2025rapidnet}. Each stage operates at a distinct spatial resolution. Downsampling layers are employed to reduce spatial dimensions by half and expand channel capacity. These downsampling layers are realized using a strided $3 \times 3$ convolution followed by Batch Normalization (Fig.\ref{fig:overall_arch} (c)) \cite{BatchNorm}. This pyramid architecture enables the learning of features across various scales, essential for robust performance on diverse visual tasks. The architecture has different model variants based on model size (VCMamba-S, VCMamba-M, and VCMamba-B), as discussed in Section \ref{ssec:network_configurations}.

Within the initial three stages, and the early parts of the fourth stage, VCMamba predominantly utilizes convolutional feed-forward network (FFN) Blocks for feature transformation and refinement. These FFN blocks are architecturally akin to the efficient inverted residual blocks prominent in mobile CNNs \cite{MobileNet, Mobilenetv2}. Structurally, each FFN block (Fig.\ref{fig:overall_arch} (d)) incorporates a core MLP module that first expands channel dimensions using a $1 \times 1$ convolution, followed by a $3 \times 3$ depthwise convolution for spatial mixing, and subsequently projects features back using another $1 \times 1$ convolution. Batch Normalization and GeLU \cite{GeLU} activation functions are applied within these MLP modules to ensure stable training and introduce non-linearity. This FFN design prioritizes efficient local feature extraction and representation learning in the higher-resolution early stages of the network.

\subsection{Multi-Directional Mamba Blocks for Global Context}
\label{ssec:mamba_blocks_detailed}
In its final stage, VCMamba transitions from convolutional FFN blocks to incorporate Multi-Directional Mamba Blocks (Fig.\ref{fig:overall_arch} (e)). This shift allows the model to leverage the rich, multi-scale local features extracted by the preceding convolutional stages and efficiently model long-range dependencies and global context across the feature maps.

Each Mamba-based block in VCMamba is a composite structure. It takes the 2D feature map from the previous layer, applies Batch Normalization, and then processes it through a core 2D-adapted Mamba module. The output of this Mamba module is combined with the input via a residual connection, followed by another Batch Normalization and a convolutional MLP, identical in structure to those used in the FFN blocks described in Section \ref{ssec:stem_and_stages}.

\begin{figure}[t]
    \centering
    \includegraphics[width=0.72\linewidth]{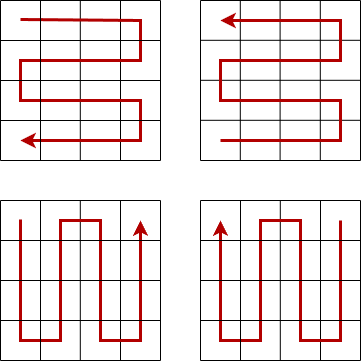} 
    \caption{The four multi-directional scanning patterns (e.g., Continuous 2D Scanning \cite{yang2024plainmamba}) employed within VCMamba's Mamba-based blocks. These patterns ensure spatial and semantic continuity when processing 2D visual tokens with 1D SSMs.}
    \label{fig:plainmamba_scan_methodology}
\end{figure}

The Mamba-based block adapts the selective scan mechanism of Mamba \cite{gu2023mamba} for 2D visual data by leveraging the scanning principles established in PlainMamba \cite{yang2024plainmamba}. Our Mamba block also leverages positional embeddings to help understand spatial relationships. As detailed in our ablation studies, the design of this block was refined to maximize performance by substituting a multiplicative branch with a skip connection, interleaving Mamba and FFN, and Layer Normalization. The process for a feature map $X \in \mathbb{R}^{B \times C \times H \times W}$ is as follows:
\begin{enumerate}

    \vspace{1mm}

    \item \textbf{Positional and Local Context Encoding:} The input feature map $X$ is first passed through a $1 \times 1$ convolution with positional embeddings to project it to an inner dimension $D_{inner}$ and normalized. The resulting features are then flattened into a sequence. To enrich these tokens with local spatial context before the main SSM operation, a $3 \times 3$ depthwise convolution followed by a SiLU activation function \cite{SiLU, GeLU} is applied. This step is crucial for preparing the visual tokens for effective sequential processing by the SSM.

    \vspace{1mm}

    \item \textbf{Multi-Directional Selective Scan:} To holistically capture 2D spatial relationships, the module employs a multi-directional scanning strategy. Instead of a single unidirectional scan, visual tokens are processed along four distinct, spatially continuous paths (e.g., row-wise and column-wise "snake" patterns), as illustrated conceptually in Figure \ref{fig:plainmamba_scan_methodology}. This approach, akin to the Continuous 2D Scanning method \cite{yang2024plainmamba}, ensures that as tokens are processed sequentially by the SSM, adjacency in the 1D sequence corresponds to spatial adjacency in the original 2D feature map, preserving crucial semantic and spatial continuity. Along each of these $k$ scan directions, the Mamba selective scan mechanism, with its input-dependent parameters ($\overline{\mathbf{A}}_{k,i}, \overline{\mathbf{B}}_{k,i}, \mathbf{C}_{k,i}, \Delta_{k,i}$), updates the hidden state $h_{k,i}$ for each token $x_i$ based on the recurrence in Eq. \eqref{eq:mamba_recurrence_prelim}.

    \vspace{1mm}

    \item \textbf{Direction-Aware Updating:} To explicitly inform the model about the nature of the 2D spatial traversal during each 1D scan, a direction-aware updating mechanism is incorporated, as introduced in \cite{yang2024plainmamba}. This involves a set of learnable parameters, $\mathbf{\Theta}_k$, each corresponding to one of the scan directions (and an initial "begin" direction). These directional parameters are integrated into the SSM's state update equation, typically by augmenting the input-dependent matrix $\overline{\mathbf{B}}_{k,i}$. The modified state update for the $k$-th scan direction and $i$-th token $x_i$ can thus be expressed as:
    \begin{equation}
        h_{k,i} = \overline{\mathbf{A}}_{k,i}h_{k,i-1} + (\overline{\mathbf{B}}_{k,i} + \overline{\mathbf{\Theta}}_{k,i})x_{i}
        \label{eq:vcmamba_direction_aware}
    \end{equation}
    where $\overline{\mathbf{\Theta}}_{k,i}$ represents the discretized directional parameter for the current token and scan path.

    \vspace{1mm}

    \item \textbf{Aggregation and Output Projection:} The feature sequences resulting from each of the four directional scans are aggregated through summation \cite{yang2024plainmamba}. This combined representation, now imbued with multi-directional context, is normalized using LayerNorm \cite{LayerNorm}. It is then followed by a $1 \times 1$ convolution and Batch Normalization.
\end{enumerate}

\noindent
This multi-directional Mamba module enables VCMamba to efficiently model global interactions and long-range dependencies within its deeper feature extraction stages.

\subsection{VCMamba Network Architecture}
\label{ssec:network_configurations}

\begin{table*}[h]
 \footnotesize
 \caption{\textbf{Architecture details of VCMamba variants (VCMamba-S, -M, -B).} Configuration of the stem, stages (number of FFN blocks and Multi-Directional Mamba (MDM) blocks), output feature map size, and channel dimensions ($C_0, C_1, C_2, C_3$).}
 \centering
 \setlength{\tabcolsep}{4pt}
 \begin{tabular}{|l|c|c|c|c|c|c|c|}
 \hline
 \multirow{2}{*}{\textbf{Stage}} & \multirow{2}{*}{\textbf{Output Size}} & \multicolumn{2}{c|}{\textbf{VCMamba-S}} & \multicolumn{2}{c|}{\textbf{VCMamba-M}} & \multicolumn{2}{c|}{\textbf{VCMamba-B}} \\ \cline{3-8}
 & & Blocks & Channels & Blocks & Channels & Blocks & Channels \\ \hline \rule{0pt}{2.5ex}
 Stem & $H/4 \times W/4$ & Conv $\times$2 & $C_0=32$ & Conv $\times$2 & $C_0=48$ & Conv $\times$2 & $C_0=64$ \\[1pt] \hline \rule{0pt}{2.5ex}
 Stage 1 & $H/4 \times W/4$ & FFN $\times$4 & $C_0=32$ & FFN $\times$4 & $C_0=48$ & FFN $\times$4 & $C_0=64$ \\[1pt]
 Downsample & $H/8 \times W/8$ & Conv & $C_1=64$ & Conv & $C_1=96$ & Conv & $C_1=128$ \\[1pt] \hline \rule{0pt}{2.5ex}
 Stage 2 & $H/8 \times W/8$ & FFN $\times$4 & $C_1=64$ & FFN $\times$4 & $C_1=96$ & FFN $\times$4 & $C_1=128$ \\[1pt]
 Downsample & $H/16 \times W/16$ & Conv & $C_2=144$ & Conv & $C_2=224$ & Conv & $C_2=320$ \\[1pt] \hline \rule{0pt}{2.5ex}
 Stage 3 & $H/16 \times W/16$ & FFN $\times$12 & $C_2=144$ & FFN $\times$12 & $C_2=224$ & FFN $\times$12 & $C_2=320$ \\[1pt]
 Downsample & $H/32 \times W/32$ & Conv & $C_3=288$ & Conv & $C_3=448$ & Conv & $C_3=512$ \\[1pt] \hline \rule{0pt}{2.5ex}
 Stage 4 & $H/32 \times W/32$ & $\begin{array}{c} \text{FFN } \times4 \\ \text{MDM } \times4 \end{array}$ & $C_3=288$ & $\begin{array}{c} \text{FFN } \times2 \\ \text{MDM } \times4 \end{array}$ & $C_3=448$ & $\begin{array}{c} \text{FFN } \times2 \\ \text{MDM } \times4 \end{array}$ & $C_3=512$ \\[1pt] \hline
 Head & $1 \times 1$ & \multicolumn{6}{c|}{Global Avg Pooling \& Linear Classifier} \\ \hline
 \end{tabular}
 \label{tab:vcmamba_arch_details}
 \end{table*}

The overall VCMamba architecture, as depicted in Figure \ref{fig:overall_arch}, integrates the convolutional stem, hierarchical stages with convolutional FFN blocks, inter-stage downsampling, and the final stage interleaving multi-directional Mamba-based blocks with convolutional FFN blocks.

We define several VCMamba variants by scaling the depth (number of blocks per stage) and width (channel dimensions), allowing for a trade-off between performance and computational cost. The configurations, VCMamba-S, VCMamba-M, and VCMamba-B, are detailed in Table \ref{tab:vcmamba_arch_details}, showing the width and depth adjusted for the smallest model size (VCMamba-S) to the largest model size (VCMamba-B). These configurations specify the number of convolutional FFN blocks and multi-directional Mamba-based blocks (MDM Blocks) in each of the four stages, along with their respective channel dimensions ($C_0, C_1, C_2, C_3$).

\section{Experiments}
\label{sec:experiments}

In this section, we detail the experimental setup and present a comprehensive performance evaluation of our proposed VCMamba architecture. We benchmark VCMamba against prominent existing architectures for image classification and semantic segmentation tasks. Our results demonstrate that VCMamba achieves excellent accuracy and computational efficiency, outperforming several state-of-the-art CNN, ViT, ViG, and other Mamba-based vision models.

\subsection{Image Classification on ImageNet-1K}

We evaluate VCMamba on the widely-used ImageNet-1K dataset \cite{imagenet1k}, which comprises approximately 1.3 million training images and 50,000 validation images across 1,000 object categories. All VCMamba models are trained from scratch for 300 epochs using a standard input resolution of $224 \times 224$. Our implementation utilizes PyTorch \cite{paszke2019pytorch} and the Timm library \cite{timm}. Following common practices for training modern vision backbones \cite{EfficientFormerv2, Deit, greedyvig2024}, our training recipe includes AdamW optimizer \cite{AdamW} with a learning rate of $2 \times 10^{-3}$ and a cosine annealing schedule, and data augmentations such as RandAugment \cite{RandAugment}, Mixup \cite{Mixup}, CutMix \cite{CutMix}, and random erasing \cite{RandomErase}.

Table \ref{tab:vcmamba_imagenet_classification} presents the ImageNet-1K top-1 accuracy and parameter counts for our VCMamba variants compared to other leading architectures, including CNNs, ViTs, and recent vision SSMs. Our largest model, \textbf{VCMamba-B}, achieves a top-1 accuracy of 82.6\% with 31.5M parameters. This performance surpasses that of PlainMamba-L3 (82.3\% with 50M parameters \cite{yang2024plainmamba}) by 0.3\% while utilizing 37\% fewer parameters, and also outperforms Vision GNN-B (82.3\% with 86.8M parameters \cite{Vision_GNN}) by 0.3\% with 64\% fewer parameters. Furthermore, \textbf{VCMamba-B} exceeds the performance of established models like ConvNeXt-T (82.1\% Acc with 29M parameters \cite{convnext}).

Our medium-sized \textbf{VCMamba-M} model, with 21.0M parameters, achieves 81.5\% top-1 accuracy, outperforming  DeiT-Small (79.9\% with 22M parameters \cite{Deit}) and PVT-Small (79.8\% with 24.5M parameters \cite{wang2021pyramid}). The lightweight \textbf{VCMamba-S} model attains 78.7\% top-1 accuracy with only 10.5M parameters, surpassing ViM-Ti (76.1\% with 7M parameters \cite{vim}) and Pyramid ViG-Ti (78.2\% with 10.7M parameters \cite{Vision_GNN}). These results underscore the effectiveness of VCMamba's hierarchical hybrid design in achieving strong classification performance across various model scales.

These results across different scales highlight the effectiveness of VCMamba's hybrid CNN-Mamba design, which leverages convolutional strengths in early stages and multi-directional Mamba capabilities in later stages, achieving a favorable accuracy-parameter trade-off.

\begin{table}[ht]
\centering
\def\arraystretch{1.1}
\caption{\textbf{ImageNet-1K classification results.} VCMamba models are compared against various CNN, ViT, ViG, and vision SSM baselines. Type indicates the primary architectural nature. Params are in millions (M).}
\label{tab:vcmamba_imagenet_classification}
\resizebox{\columnwidth}{!}{
\begin{tabular}{lcccc}
\toprule
\textbf{Model} & \textbf{Type} & \textbf{Params (M)} & \textbf{GMACs} & \textbf{Top-1 (\%)} \\
\midrule
ResNet-152 \cite{resnet} & CNN & 60.2 & 11.5 & 81.8 \\
ConvNeXt-T \cite{convnext} & CNN & 29.0 & 4.5 & 82.1 \\
ConvNeXt-S \cite{convnext} & CNN & 50.0 & 8.7 & 83.1 \\
\midrule
DeiT-Small \cite{Deit} & ViT & 22.0 & 4.6 & 79.9 \\
DeiT-Base \cite{Deit} & ViT & 86.0 & 16.8 & 81.8 \\
PVT-Tiny \cite{wang2021pyramid} & ViT & 13.2 & 2.0 & 75.1 \\
PVT-Small \cite{wang2021pyramid} & ViT & 24.5 & 3.8 & 79.8 \\
EfficientFormer-L1 \cite{EfficientFormer} & CNN-ViT & 12.3 & 1.3 & 79.2 \\
EfficientFormer-L3 \cite{EfficientFormer} & CNN-ViT & 31.3 & 3.9 & 82.4 \\
EfficientFormer-L7 \cite{EfficientFormer} & CNN-ViT & 82.1 & 10.2 & 83.3 \\
PoolFormer-S12 \cite{MetaFormer} & CNN-Pooling & 12.0 & 1.8 & 77.2 \\
PoolFormer-S24 \cite{MetaFormer} & CNN-Pooling & 21.0 & 3.4 & 80.3 \\
PoolFormer-S36 \cite{MetaFormer} & CNN-Pooling & 31.0 & 5.0 & 81.4 \\
PoolFormer-M36 \cite{MetaFormer} & CNN-Pooling & 56.0 & 8.8 & 82.1 \\
\midrule
ViG-S \cite{Vision_GNN} & GNN & 22.7 & 4.5 & 80.4 \\
ViG-B \cite{Vision_GNN} & GNN & 86.8 & 17.7 & 82.3 \\
Pyramid ViG-Ti \cite{Vision_GNN} & GNN & 10.7 & 1.7 & 78.2 \\
Pyramid ViG-S \cite{Vision_GNN} & GNN & 27.3 & 4.6 & 82.1 \\
Pyramid ViG-M \cite{Vision_GNN} & GNN & 51.7 & 8.9 & 83.1 \\
Pyramid ViG-B \cite{Vision_GNN} & GNN & 92.6 & 16.8 & 83.7 \\
\midrule
ViM-Ti \cite{vim} & SSM & 7.0 & - & 76.1 \\
ViM-S \cite{vim} & SSM & 26.0 & - & 80.5 \\
PlainMamba-L1 \cite{yang2024plainmamba} & SSM & 7.0 & 3.0 & 77.9 \\
PlainMamba-L2 \cite{yang2024plainmamba} & SSM & 25.0 & 8.1 & 81.6 \\
PlainMamba-L3 \cite{yang2024plainmamba} & SSM & 50.0 & 14.4 & 82.3 \\
MambaVision-T \cite{hatamizadeh2024mambavision} & SSM-ViT & 31.8 & 4.4 & 82.3 \\
\midrule
\textbf{VCMamba-S (Ours)} & \textbf{CNN-SSM} & \textbf{10.5} & \textbf{1.1} & \textbf{78.7} \\
\textbf{VCMamba-M (Ours)} & \textbf{CNN-SSM} & \textbf{21.0} & \textbf{2.3} & \textbf{81.5} \\
\textbf{VCMamba-B (Ours)} & \textbf{CNN-SSM} & \textbf{31.5} & \textbf{4.0} & \textbf{82.6} \\
\bottomrule
\end{tabular}
 }
\end{table}

\subsection{Semantic Segmentation on ADE20K}

To evaluate VCMamba's capabilities on dense prediction tasks, we conduct semantic segmentation experiments on the ADE20K dataset \cite{ADE20K}. ADE20K contains 20K training images and 2K validation images, encompassing 150 semantic categories \cite{EfficientFormer, munir2025rapidnet}. We build VCMamba with Semantic FPN \cite{kirillov2019panoptic} as the segmentation decoder, following established methodologies \cite{MetaFormer, EfficientFormer, MobileViG, greedyvig2024}. The VCMamba backbones are initialized with their ImageNet-1K pre-trained weights. The models are then fine-tuned for 40K iterations. We use the AdamW optimizer \cite{AdamW} with an initial learning rate of $2 \times 10^{-4}$, which is decayed using a polynomial schedule with a power of 0.9. The training input resolution is $512 \times 512$ \cite{EfficientFormer, MobileViG, greedyvig2024}.

\begin{table}[ht]
\centering
\def\arraystretch{1.05}
\footnotesize
\caption{\textbf{Semantic segmentation results on ADE20K}. Parameters are backbone-only. Bold entries indicate results obtained using VCMamba.}
\label{tab:vcmamba_ade20k_segmentation}
\begin{tabular}[t]{|c|c|c|c|c|}
\hline
{\textbf{Backbone}} & {\textbf{Params (M)}} & {\textbf{mIoU}} \\ \hline
ResNet18 \cite{resnet} & 11.7 & 32.9 \\ \hline
PoolFormer-S12 \cite{MetaFormer} & 12.0 & 37.2 \\ \hline
EfficientFormer-L1 \cite{EfficientFormer} & 12.3 & 38.9 \\ \hline
FastViT-SA12 \cite{FastViT} & 10.9 & 38.0 \\ \hline
\textbf{VCMamba-S (Ours)} & \textbf{10.5} & \textbf{42.0}  \\ \hlineB{5}
ResNet50 \cite{resnet} & 25.5 & 36.7 \\ \hline
PVT-Small \cite{wang2021pyramid} & 24.5 & 39.8 \\ \hline
PoolFormer-S36 \cite{MetaFormer} & 31.0 & 42.0 \\ \hline
FastViT-SA36 \cite{FastViT} & 30.4 & 42.9 \\ \hline
RapidNet-B \cite{munir2025rapidnet} & 30.5 & 43.8 \\ \hline
EfficientFormer-L7 \cite{EfficientFormer} & 82.1 & 45.1 \\ \hline
\textbf{VCMamba-B (Ours)} & \textbf{31.5} & \textbf{47.1} \\ \hlineB{5}
\end{tabular}
\end{table}

As shown in Table \ref{tab:vcmamba_ade20k_segmentation}, VCMamba demonstrates strong performance in semantic segmentation. Our \textbf{VCMamba-S} (10.5M backbone parameters) achieves 42.0 mIoU, outperforming other lightweight models such as EfficientFormer-L1 (38.9 mIoU \cite{EfficientFormer}). Our larger model, \textbf{VCMamba-B} (31.5M backbone parameters), achieves an impressive 47.1 mIoU. This surpasses FastViT-SA36 by 4.2 mIoU and the much larger EfficientFormer-L7 (45.1 mIoU with an 82.1M backbone \cite{EfficientFormer}) by 2.0 mIoU, despite \textbf{VCMamba-B} utilizing approximately 62\% fewer parameters in its backbone. These results highlight the efficacy of VCMamba's architecture in learning powerful representations for dense prediction tasks, effectively leveraging its hybrid convolutional and multi-directional Mamba design.

\subsection {Ablation Studies}
To validate our architectural design choices, we conduct a series of ablation studies on ImageNet-1K, starting from a baseline of using PlainMamba \cite{yang2024plainmamba} layers for our final stage and progressively integrating key components of our proposed VCMamba. This baseline model achieves 80.2\% top-1 accuracy with 33.0 M parameters. The step-by-step improvements are detailed in Table \ref{tab:vcmamba_ablation}.

First, we refine the internal structure of the Mamba block by substituting a multiplicative branch with a simple skip connection, which improves accuracy to 80.7\% while slightly reducing parameters to 31.0 M. Next, to better fuse feature representations, we interleave Mamba and FFN layers followed by batch normalization, further boosting performance to 81.5\%. We then observe that adding an additional LayerNorm \cite{LayerNorm} within the Mamba layer provides a significant gain, reaching 82.2\% accuracy. Next, we replaced the linear layer heads with convolutions resulting in an accuracy of 82.5\%. Finally, to ensure stable feature distributions across the model's hierarchy, we wrap each stage with batch normalization layers. This final model is our \textbf{VCMamba-B}, which achieves a top-1 accuracy of 82.6\%, a total improvement of 2.4\% over our CNN-Mamba baseline. This series of ablations validates the efficacy of the final VCMamba architecture.

\begin{table}[h]
\centering
\def\arraystretch{1.0}
\footnotesize
\caption{\textbf{Ablation study on the VCMamba-B architecture.} We start with our CNN-Mamba baseline and incrementally add modifications. Acc. denotes ImageNet-1K Top-1 accuracy.}
\label{tab:vcmamba_ablation}
\begin{tabular}{l c c c}
\toprule
\textbf{Model / Configuration} & \textbf{Params (M)} & \textbf{Acc. (\%)}  \\
\midrule
1. Baseline (PlainMamba Stage 4) & 33.0 & 80.2  \\
2. + Skip Connection & 31.0 & 80.7  \\
3. + Interleaved Mamba \& FFN & 31.1 & 81.5 \\
4. + LayerNorm & 31.1 & 82.2 &  \\
5. + Replace Linear Projection & 31.4 & 82.5 & \\
6. \textbf{VCMamba-B} (Stage-wise BN) & \textbf{31.5} & \textbf{82.6}  \\
\bottomrule
\end{tabular}
\end{table}

\section{Conclusion}
\label{sec:conclusion}
In this paper, we have introduced VCMamba, a novel hierarchical vision backbone designed to synergistically bridge the robust local feature extraction capabilities of Convolutional Neural Networks (CNNs) with the efficient global context modeling of State Space Models (SSMs). VCMamba employs a multi-stage architecture, utilizing convolutional blocks in its early and intermediate stages to build rich, multi-scale feature hierarchies. In its deeper, lower-resolution stages, it transitions to multi-directional Mamba blocks to effectively capture long-range dependencies. This hybrid design allows VCMamba to maintain linear complexity in its Mamba stages, offering a scalable solution for high-resolution visual understanding.

Our extensive evaluations demonstrate VCMamba's strong performance and efficiency across ImageNet-1K classification and ADE20K semantic segmentation. Notably, VCMamba-B achieves 82.6\% top-1 accuracy on ImageNet-1K, surpassing models like PlainMamba-L3 by 0.3\% with 37\% fewer parameters, and obtains 47.1 mIoU on ADE20K, exceeding EfficientFormer-L7 by 2.0 mIoU while using 62\% fewer parameters. These results validate VCMamba as a compelling and efficient backbone for a diverse range of computer vision tasks.

\section{Acknowledgements}
\label{Sec:Acknowledgement}

This work is supported in part by the NSF grant CNS 2007284, the iMAGiNE Consortium (\url{https://imagine.utexas.edu/}), and a UT Cockrell School of Engineering Doctoral Fellowship.


{
    \small
    \bibliographystyle{ieeenat_fullname}
    \bibliography{main}
}

\end{document}